\pdfoutput=1
\documentclass[11pt]{article}

\usepackage{ACL2023}

\usepackage{times}
\usepackage{latexsym}
\usepackage{paralist}

\usepackage[T1]{fontenc}
\usepackage[utf8]{inputenc}
\usepackage[russian,english]{babel}

\usepackage{microtype}

\usepackage{inconsolata}

\usepackage{booktabs}
\usepackage{graphicx}
\usepackage{amsmath}
\usepackage{tcolorbox}
\usepackage{vcell}
\usepackage{soul}
\usepackage{xcolor} 

\sethlcolor{pink}

\newcommand{\multirowcell}[1]{\begin{tabular}[c]{@{}c@{}}#1\end{tabular}}

\let\svthefootnote\thefootnote
\newcommand\freefootnote[1]{%
  \let\thefootnote\relax%
  \footnotetext{#1}%
  \let\thefootnote\svthefootnote%
}

\title{
Vikhr: The Family of Open-Source Instruction-Tuned Large Language Models for Russian
}

\author{Aleksandr Nikolich\textsuperscript{$\diamondsuit$}  \\
  ITMO University \\
  \texttt{alexdragannikolich@gmail.com}\And
  Konstantin Korolev\textsuperscript{$\diamondsuit$} \\
  HSE University \\
  \texttt{korolevko@icloud.com}\And Sergei Bratchikov \\
  Vikhr\\ 
  \texttt{hivaze.me@gmail.com}\AND
  Igor Kiselev \\
  Accenture \\
  \texttt{igor.kiselev@accenture.com}\And
  Artem Shelmanov \\
  MBZUAI \\
  \texttt{artem.shelmanov@mbzuai.ac.ae}
}

\begin{document}
\maketitle

\freefootnote{$\diamondsuit$ Equal contribution}

\begin{abstract}
There has been a surge in the development of various Large Language Models (LLMs). However, text generation for languages other than English often faces significant challenges, including poor generation quality and reduced computational performance due to the disproportionate representation of tokens in the model's vocabulary. In this work, we address these issues by developing a pipeline for the adaptation of English-oriented pre-trained models to other languages and constructing efficient bilingual LLMs. Using this pipeline, we construct Vikhr, a series of bilingual open-source instruction-following LLMs designed specifically for the Russian language. ``Vikhr'' refers to the name of the Mistral LLM series and means a ``strong gust of wind.''
Unlike previous Russian-language models that typically rely on 
LoRA adapters on top of English-oriented models, sacrificing performance for lower training costs, Vikhr features an adapted tokenizer vocabulary and undergoes continued pre-training and instruction tuning of all weights. This not only enhances the model's performance but also significantly improves its computational and contextual efficiency.
We also expanded the instruction datasets and corpora for continued pre-training. 
The model weights, instruction sets, and code are publicly available.\footnote{\url{https://huggingface.co/Vikhrmodels}}
\end{abstract}

\section{Introduction}

Instruction tuning has unlocked in Large Language Models (LLMs) vast zero-shot capabilities without the need for careful prompt engineering \cite{ouyang2022training}. The most rapid research and development efforts are currently devoted to English LLMs. 
There has been a surge in English open-source models: Llama series \cite{touvron2023llama,touvron2023llama2}, Mistral series \cite{jiang2023mistral}, Vicuna series \cite{vicuna2023}, etc. This growth is driven by the abundance of raw training data in English and dedicated efforts to create extensive sets of instruction-output pairs. Even though LLMs oriented on English have some multilingual capabilities \cite{zhao2024llama} due to the presence of small amounts of text in various languages within their training datasets \cite{touvron2023llama}, their overall performance in non-English languages remains relatively limited. Although they can usually generate portions of coherent texts, these models struggle with reasoning in non-English languages, lack culture-specific knowledge, and are highly inefficient in terms of tokenization. This inefficiency stems from how byte-pair tokenization algorithms operate, as they break down infrequent words into multiple tokens. Since multilingual data typically represents a small portion of the training dataset, non-English words are often split into many pieces. As a result, this increases the number of steps during prompt processing and text generation, reduces the effective context window, and ultimately degrades overall performance \cite{tikhomirov2023impact,petrov2024language}. This disparity places non-English languages at a disadvantage.

There is also a research direction focused on developing multilingual LLMs designed to perform well across multiple popular languages:  BLOOMz \cite{muennighoff2023crosslingual}, mGPT \cite{shliazhko2022mgpt}, Bactrian-X \cite{li2023bactrian}, PALO \cite{maaz2024palo}, Aya101 from CohereAI \cite{ustun2024aya}, etc. 
These models are typically trained on rich multilingual datasets and are less skewed towards English. 
However, when aiming to perform well across multiple languages simultaneously, these models must still share their vocabulary and parameters. This often hinders their performance for each particular language in isolation, especially for the popular smaller model sizes, such as 7B and 13B.

The aim of maximizing the LLM performance for a specific language within a certain number of parameters has led researchers to develop bilingual LLMs \cite{sengupta2023jais,pieri2024bimedix,faysse2024croissantllm}. 
These LLMs prioritize a regional language, e.g. Jais \cite{sengupta2023jais} focuses on Arabic, but they are trained also on English data.
The inclusion of English data in pre-training alongside regional language data is motivated by the significantly larger volume of English data available. This helps LLMs substantially enhance skills such as logical and common sense reasoning, which are also applied when generating text in a regional language. Bilingual LLMs are a promising direction as they can remain small and efficient, but at the same time comprehensively capture linguistic nuances and cultural contexts of the regional language.

This work seeks to develop a pipeline for adapting English LLMs to other languages, facilitating the development of bilingual LLMs. Specifically, we aim to build an instruction-following bilingual LLM for Russian and English that could be used for multilingual natural language processing research.

Russian is one of the high-resource languages and is typically represented in multilingual LLMs. Additionally, there are several proprietary closed-source LLMs, such as MTS AI, GigaChat, and YandexGPT, that meet or even surpass their English-oriented flagship competitors when it comes to text processing and generation in Russian. However, controllable research often requires white-box access to LLM logits and layer outputs, the ability to modify weights and model architecture, and consistent answers for reproducibility, which is often impossible in closed-source LLMs due to their constant development and retirement. There are only a few open-source LLMs designed for Russian: Saiga \cite{gusev_rulm_2023}, ruGPT \cite{aiforever_rugpts_2024}, ruadapt \cite{tikhomirov2023impact}, and some others. Of these, only Saiga and ruadapt are instruction-tuned. We aim to fill the lack of instruction-tuned open-source LLM for Russian that is both efficient and effective.

Building even a small LLM tailored to a specific language from scratch demands a lot of computational resources. Consequently, many researchers opt to fine-tune LoRA adapters \cite{hu2021lora} for English-oriented LLMs using some language-specific data. While this approach can improve model generation quality, it does not address computational inefficiency because the tokenizer and model vocabulary remain unchanged. In contrast, our approach not only fine-tunes a base LLM on Russian language data but also reconstructs its underlying tokenizer and vocabulary, alongside suggesting an improved method for continued pre-training. Additionally, we have significantly expanded the available Russian datasets for instruction tuning.

Contributions of the paper are the following:
\begin{compactitem}
    \item We have developed a pipeline for adapting English-oriented LLMs to other languages. The pipeline implements vocabulary adaptation, continued pre-training with regularization to prevent ``catastrophic forgetting,'' and instruction tuning.

    \item Using the pipeline, we have constructed Vikhr
    -- an open-source instruction-following LLM oriented on the Russian language. In addition to its high generation quality, Vikhr features an efficient tokenizer that enables rapid text generation and good context utilization.

    \item We have expanded the datasets for continued pre-training of Russian language models and previously available instruction datasets.

    \item We have constructed two evaluation benchmarks for Russian LLMs by translating the English MMLU \cite{hendrycksmeasuring} and MMLU-pro \cite{wang2024mmlupro} benchmarks.
    
    \item We conducted ablation studies of various pipeline components.
    The studies confirm the effectiveness and validity of the individual components within our LLM adaptation pipeline.

\end{compactitem}

\section{Related Work}
\label{sec:related_work}

One of the first prominent series of generative LLMs for Russian is ruGPT \cite{aiforever_rugpts_2024,zmitrovich2023family}. The authors developed several models, trained on the standard language modeling task, with sizes reaching up to 13 billion parameters. These models were created from scratch and trained on a large Russian corpus, enabling them to capture the linguistic nuances of Russian more effectively than multilingual models. Additionally, since the training data was mostly in Russian, these models also have efficient tokenization. However, the lack of multilingual data (e.g. in English) limits their performance. Notably, the ruGPT models are not instruction-tuned.

\citet{gusev_rulm_2023} suggests to leverage the reasoning capabilities of existing English-oriented LLMs and adapt them to the Russian language by training LoRA adapters. They created an Alpaca-like set of Russian instruction-output pairs and performed instruction tuning on it. As a result, they have established a series of models called Saiga, which demonstrate competitive performance and have been a reasonable choice for off-the-shelf open-source Russian LLMs for the past years. However, the tokenizer in these models is not adapted, so they experience issues with context size and computational  efficiency.

\citet{tikhomirov2023impact} address these issues in Saiga. In addition to model tuning on Russian data, they also adapted the model tokenizer. They note that improving tokenization helps to both enhance the efficiency of the model and its performance while reducing memory consumption. However, during continued pre-training, the authors froze the model weights, except for LM heads and token embeddings, which leads to suboptimal performance.

In this work, we take advantage of pre-trained English-oriented LLMs, adapt LLM tokenizer for better computational and contextual efficiency, leverage continued pre-training on vast Russian-language corpora with regularization for preventing ``catastrophic forgetting'', construct a novel extended set of Russian instruction-output pairs, and perform instruction tuning. The created LLM adaptation pipeline along with the data for continued pre-training and instruction tuning enables Vikhr to achieve good generation quality for Russian and demonstrate high computational efficiency.

\section{LLM Construction Pipeline}

The construction of Vikhr starts from one of the English-oriented LLMs. In this work, we discuss the Vikhr model based on Mistral 7B \cite{jiang2023mistral}. The strong logical and common-sense reasoning capabilities, as well as the extensive world knowledge present in Mistral LLMs, provide an excellent starting point for our model. These features partially transfer to Vikhr, enhancing its performance in generating text in Russian. The process of the LLM adaptation to Russian starts with the vocabulary adaptation. Then we perform continued pre-training of the LLM on large Russian datasets to mitigate the vocabulary shift and introduce culture-specific knowledge. Finally, we perform fine-tuning of Vikhr on a set of instruction-output pairs in Russian.

\subsection{Vocabulary Adaptation}

\begin{table}[t]
\centering
\footnotesize
\begin{tabular}{p{0.2\linewidth} | p{0.15\linewidth} | p{0.5\linewidth}}
\toprule
\textbf{Content} & \textbf{Length} & \textbf{Tokenization Result} \\
\midrule
Original Sentence & 31 & \textcyrillic{(ru) Машинное обучение изменяет мир \newline [(en) Machine learning changes the world.} \\
\midrule
Mistral Tokenizer & 13 & [\textcyrillic{`Ма',
 `шин',
 `ное',
 `об',
 `у',
 `чение',
 `из',
 `мен',
 `я',
 `ет',
 `ми', i
 `р' }] \\
\midrule
Vikhr Tokenizer & 7 & [\textcyrillic{`Ма', `шин', `ное', `обучение', `изменяет', `мир'}] \\
\bottomrule
\end{tabular}
\caption{A comparison of tokenization between the original Mistral model and Vikhr.}
\label{tab:vocab_example}
\end{table}

\begin{figure}[t]
    \centering
    \includegraphics[width=\linewidth]{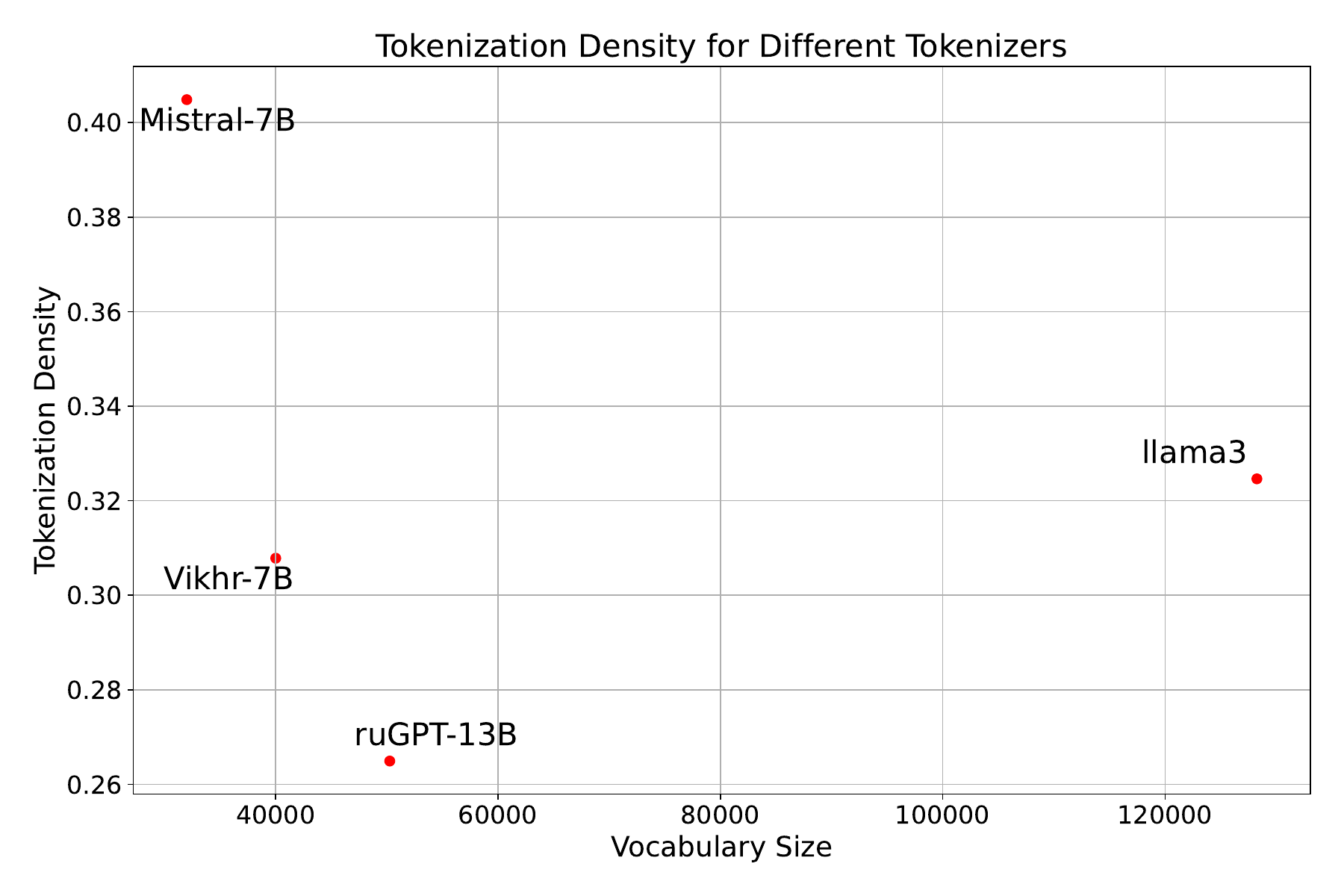}
    \caption{The efficiency of the Vikhr tokenizer for Russian in comparison to tokenizers of other models.}
    \label{fig:tokneizer}
\end{figure}

The big drawback of English-oriented LLMs is that each Russian word would be split into multiple tokens: a common case is when symbols in the word become individual tokens (see example in Table \ref{tab:vocab_example}). This slows down the generation by multiple times, reduces the amount of information that could be stored in the context, and drastically hurts the generation quality. 

To mitigate this problem in Vikhr, we adopt the approach suggested in \cite{chinese-llama-alpaca,tikhomirov2023impact}, where authors rebuild the tokenizer using a language-specific corpus.
In particular, we trained a SentencePiece tokenizer \cite{kudo2018sentencepiece} with a 40k vocabulary on the RuLM dataset \cite{gusev_rulm_2023}.
As can be seen from Figure \ref{fig:tokneizer}, the resulting tokenizer for Russian is much more efficient than the tokenizer of the original English-oriented model.

\begin{table}[t]
    \centering
    \footnotesize
    \begin{tabular}{l|c|c}
    \toprule 
         \textbf{Data Source} & \multirowcell{\textbf{Approx. size} \\ \textbf{(GB)}} & \multirowcell{\textbf{Tokens} \\ \textbf{(Billion)}}\\
         \midrule
Scientific papers & 20 & 2.5\\
News articles & 4 & 1 \\
Wikipedia & 25 & 4 \\
Habr & 6 & 1\\
Other sources & 20 & 2.5\\
\bottomrule
    \end{tabular}
    \caption{The statistics of the Russian-language datasets for continued pre-training.}
    \label{tab:cp_statistics}
\end{table}

\subsection{Continued Pre-training}

The new vocabulary requires new embedding matrices and LM heads. The tokens that were present in the original vocabulary are initialized with the old embeddings; the new tokens are initialized by averaging the embeddings of their pieces in the original embedding matrix \cite{hewitt2021initializing}. 
The similar approach is also applied to LM heads. Training models with these modifications demands much more computational resources than the mainstream method of adapting LLMs to new languages using LoRA adapters \cite{hu2021lora}. This is because  it involves continued pre-training of the entire model and requires much more language-specific data to mitigate the shift in the vocabulary.

The dataset for continued pre-training is constructed from high-quality sources, including Russian Wikipedia, news articles, scientific papers from peer-reviewed journals and conferences, and top 100k up-voted posts on Habr -- a popular online blog community focused on technology, software development, and science. The statistics of these datasets are presented in Table \ref{tab:cp_statistics}. 
We performed deduplication of the collection on the level of paragraphs using the MIHash algorithm \cite{mihash}. Furthermore, we performed filtration of the collected data. For this purpose, we annotated 20k documents using GPT-4-turbo with the aim to assess their informativeness, usefulness for studying, grammatical correctness, style, and safety. Using these annotations, we fine-tuned a RuBERT-tiny\cite{dale_tiny_and_fast_bert_2021} filtration model and applied it to the deduplicated corpus. After filtration, the total number of tokens left for continued pre-training is 11 billion.

We observed that continued pre-training of a LLM can partially diminish the reasoning capabilities present in the original English-oriented model, significantly impacting overall performance. 
In our preliminary experiments, a model that underwent continued pre-training may demonstrate even worse performance on Russian benchmarks than the original English-oriented model. To alleviate this ``catastrophic forgetting'' in reasoning, we use the loss regularization with the KL penalty between the probability distribution of Vikhr and the reference English-oriented LLM:   
\begin{equation}
L_{\text{Vikhr}}=L_{\text{CE}}+K L\left(P_{\text{Vikhr}} \| P_{\text {Ref }}\right).
\label{eq:loss}
\end{equation}
In practice, we implement the regularization using the SLERP interpolation of model losses \cite{goddard2024arcee}. 

\begin{table}
    \centering
    \footnotesize
    \begin{tabular}{l|c}
    \toprule 
         \textbf{Hyperparam.} & \textbf{Value}\\
         \midrule
         LR & $1 \times 10^{-3}$ \\
AdamW eps & $1 \times 10^{-8}$ \\
Num warmup steps & 10 \\
AdamW betas & $0.99$, $0.95$ \\
Accumulation steps & $128 $\\
Batch size & $3 $\\
Epochs & $1 $\\
Sequence length & $1024 $\\
\bottomrule
         
    \end{tabular}
    \caption{The hyperparameters for continued pre-training.}
    \label{tab:hyper_pretrain}
\end{table}

To speed up the process of continued pre-training, we use an optimized Flash attention implementation.\footnote{\url{https://huggingface.co/docs/optimum/bettertransformer/tutorials/convert}} 
As an optimization algorithm, we leverage AdamW, as it trades some memory efficiency in favor of robustness to the hyperparameter choice. The hyperparameters used for continued pre-training are presented in Table \ref{tab:hyper_pretrain}.

\subsection{Instruction Tuning}

Instruction tuning is an essential step in reaching high zero-shot performance with LLMs. It also allows obtaining a more natural communication with the model without complex prompting. Further fine-tuning techniques such as RLHF \cite{ouyang2022training} or DPO \cite{rafailov2024direct}, which require input from the assessors, are also crucial for such tasks as multicriteria alignment. However, the most significant  performance gains are still achieved through instruction tuning \cite{jha2023limit}. 

Previously, \citet{gusev_rulm_2023} constructed an open-source set of instruction-output pairs for the Russian language (Saiga). The core Saiga dataset was created similar to Alpaca by querying ChatGPT (gpt-3.5-turbo) \cite{taori2023stanford}. In this work, we extend this set by translating two English instruction datasets. First, we translated instructions for the FLAN model \cite{wei2021finetuned} and generated answers in Russian using ChatGPT. Originally, FLAN instructions were constructed automatically from annotated datasets using templates to facilitate multitask and zero-shot capabilities of seq2seq models. Later, it was shown that this data also helps to improve decoder-only chat-oriented models as well. Second, we construct Veles\footnote{\url{ https://huggingface.co/datasets/Vikhrmodels/Veles-2.5}} by translating the English OpenHermes \cite{OpenHermes2.5} instruction dataset.
Third, we incorporate without translation Nectar\footnote{\url{https://huggingface.co/datasets/berkeley-nest/Nectar}} \cite{starling2023} -- an English instruction dataset. This ensures that Vikhr maintains strong performance in English as well.

Similar to the corpus for continued pre-training, we performed deduplication of the instruction set. 
Since the majority of the outputs were machine generated, there are many low-quality outputs. To mitigate this problem, we filtered out low-quality pairs using a reward model trained on human data.
For the reward model, we selected the multilingual-e5-large model \cite{wang2024multilingual}. This model was particularly suitable for our needs due to its ability to handle multilingual data efficiently, ensuring that the classifier could accurately assess the quality of responses in both Russian and English. We trained the reward model on the answer preference dataset\footnote{\url{https://huggingface.co/datasets/Vikhrmodels/sbs}}, which was collected from human-written prompts and annotated using GPT-4.
By applying this reward model, we filtered out low-quality instruction-output pairs, significantly enhancing the overall performance and reliability of our instruction datasets. The statistics of the Vikhr instruction datasets are presented in Table \ref{tab:stat_instr}.

\begin{table}
    \centering
    \footnotesize
    \begin{tabular}{l|c|c}
    \toprule
         \textbf{Instruction Set} & \textbf{Language} & \textbf{\# instances}\\
         \midrule
         Veles & Russian & 30k \\ 
         Nectar & English & 50k \\ 
         Saiga & Russian & 100k \\ 
         ruFLAN & Russian & 500k \\
         \bottomrule
    \end{tabular}
    \caption{The statistics of the instruction datasets.}
    \label{tab:stat_instr}
\end{table}

\begin{table}
    \centering
    \footnotesize
    \begin{tabular}{l|c}
    \toprule 
         \textbf{Hyperparam.} & \textbf{Value}\\
         \midrule
         LR & $1 \times 10^{-5}$ \\
AdamW, eps & $1 \times 10^{-8}$ \\
Num warmup steps & 10 \\
AdamW, betas & $0.99$, $0.95$ \\
Accumulation steps & $64 $\\
Batch size & $3 $\\
Num epochs & $3 $\\
Sequence length & $1024 $\\
\bottomrule
         
    \end{tabular}
    \caption{The hyperparameters for instruction tuning.}
    \label{tab:hyper_instrtune}
\end{table}

\begin{table*}[ht]
\centering
\tiny
\begin{tabular}{l|l|l}
\textbf{Prompt} & \textbf{Answer of base Mistral-0.2-Instruct} & \textbf{Answer of Vikhr} \\
\toprule

\begin{tcolorbox}[size=fbox,
		title=,
		colback=gray!2!white,
		colframe=white!15!white,
            coltitle=black,
		width=16em,
		sharp corners=northwest,
		]
\tiny
\textcyrillic{Что такое чебурек? \newline [What does cheburek means?]}
\end{tcolorbox}

&

\begin{tcolorbox}[size=fbox,
		title=,
		colback=gray!2!white,
		colframe=white!15!white,
            coltitle=black,
		width=27em,
		sharp corners=northwest,
		]
\tiny
\textcyrillic{Чебурек - это классическое \hl{dish} среднеазиатской и русской кухни, 
которое включает в себя тонкий \hl{lavash}, завернутый в форме полумесяца или кольца.
Он обычно наполнен \hl{meat} (говядиной, бараниной или курицей) и \hl{cheese}.
История чебурека начинается с древних тюркских племен, которые готовили подобные \hl{dishes} из мяса и теста. ... 
Чебурек часто подается как горячее \hl{dish}, жареное на сковороде или в духовке.  Тесто обычно тонкое и хрустящее снаружи, а внутри мягкое и сочное. Мясо и сыр добавляют \hl{flavor depth} и питательную ценность.}
\end{tcolorbox}

& 

\begin{tcolorbox}[size=fbox,
		title=,
		colback=gray!2!white,
		colframe=white!15!white,
            coltitle=black,
		width=27em,
		sharp corners=northwest,
		]
\tiny
\textcyrillic{
Чебурек — это традиционное блюдо кавказской кухни, которое представляет собой большой, тонкий пирог из пресного теста с мясной начинкой, обжаренный во фритюре. Внутри чебурека обычно находится сочный фарш из баранины или говядины, приправленный луком, солью, перцем и другими специями. Благодаря обжарке в масле, тесто становится хрустящим снаружи, а начинка остается мягкой и сочной.
История чебурека уходит корнями в древние времена, когда кочевники готовили его на раскалённых камнях или в больших котлах с горячим маслом. 
}
\end{tcolorbox} \\
\midrule

\begin{tcolorbox}[size=fbox,
		title=,
		colback=gray!2!white,
		colframe=white!15!white,
            coltitle=black,
		width=16em,
		sharp corners=northwest,
		]
\tiny
\textcyrillic{Расскажи на русском языке что такое хачапури  \newline
[Tell in Russian what hatchapuri is]}
\end{tcolorbox}

&

\begin{tcolorbox}[size=fbox,
		title=,
		colback=gray!2!white,
		colframe=white!15!white,
            coltitle=black,
		width=27em,
		sharp corners=northwest,
		]
\tiny
\textcyrillic{
 Хачапури (хачпури, хачъапури) – это традиционное блюдо грузинской кухни. Это тип \hl{пицце} или \hl{за Francia}, выполненной в виде круглой или полукруглой лепёшки, покрытой верху \hl{layerм layerim} – смесью тонко \hl{нарезанныхimo} \hl{эplesses}, лука .. 
 }
\end{tcolorbox}

& 

\begin{tcolorbox}[size=fbox,
		title=,
		colback=gray!2!white,
		colframe=white!15!white,
            coltitle=black,
		width=27em,
		sharp corners=northwest,
		]
\tiny
\textcyrillic{
Хачапури — это традиционное грузинское блюдо, представляющее собой лепешку с сыром внутри. Вот рецепт хачапури. Ингредиенты:
- Для теста:
  - 3 стакана муки
  - 1 ч. л. соли
  - 1 ч. л. сахара
    - 100 мл теплой воды
  - Для смазывания: 1 ст. л. топленого масла
 ...
}
\end{tcolorbox}

\\
\midrule

\begin{tcolorbox}[size=fbox,
		title=,
		colback=gray!2!white,
		colframe=white!15!white,
            coltitle=black,
		width=16em,
		sharp corners=northwest,
		]
\textcyrillic{Расскажи на русском языке как приготовить уху \newline [Tell in Russian how to cook a fish soup]}
\end{tcolorbox}

&

\begin{tcolorbox}[size=fbox,
		title=,
		colback=gray!2!white,
		colframe=white!15!white,
            coltitle=black,
		width=27em,
		sharp corners=northwest,
		]
\tiny
\textcyrillic{Уха (из Рыбы) – традиционное блюдо русской кухни, которое готовится \hl{predominantly} из щуки или окуня. \hl{Рас**те} процедуру приготовления ухи в домашних условиях:
Выберите чистую рыбу. Очистить её от \hl{шерсти} \hl{иICScoне} всех мелких костей s...} 
\end{tcolorbox}

&

\begin{tcolorbox}[size=fbox,
		title=,
		colback=gray!2!white,
		colframe=white!15!white,
            coltitle=black,
		width=27em,
		sharp corners=northwest,
		]
\tiny
\textcyrillic{
Уха — это традиционное русское блюдо, представляющее собой наваристый суп из рыбы, который варится на медленном огне. Вот базовый рецепт ухи, который можно адаптировать под свои вкусовые предпочтения.
Ингредиенты:
1 кг свежей рыбы (можно использовать несколько видов, например, карп, щука, судак)
3-4 картофелины
1 крупная луковица
2-3 моркови ... 
}
\end{tcolorbox}

\\

\bottomrule
\end{tabular}
    
\caption{Comparison of texts generated by the base LLM Mistral-0.2-Instruct (on the left) with texts generated by our model Vikhr (on the right). Issues in the output are highlighted with red.
}
\label{tab:qualitative}
\end{table*}

Contrary to Saiga, we do not use LoRA adapters, and just as in the phase of continued pre-training, we update all model parameters. The hyperparameters for the instruction tuning phase are presented in Table \ref{tab:hyper_instrtune}.

\subsection{Hardware}
Vikhr was trained on eight NVIDIA A100 GPUs 80GB. We spent approximately 1,000 GPU hours for the continued pre-training phase and 60 hours for instruction tuning.

\section{Experiments}

\subsection{Experimental Setup}

For evaluation, we use ruXNLI~\cite{conneau2018xnli}, and we also constructed Ru-MMLU and Ru-MMLU-pro by automatically translating the original English MMLU and MMLU-pro to Russian. Translation was done by GPT-3.5 and GPT-4, respectively.
We report the accuracy@1 score.

\begin{table}[t]
    \centering
    \footnotesize
    \begin{tabular}{c|cc}
        \toprule
        \textbf{Vocab. Size} & \textbf{PPL$\downarrow$} & \textbf{ruXNLI$\uparrow$} \\
        \midrule
        33k & \textbf{10.2}  & 0.42  \\
        \underline{40k} & 14.4  & \textbf{0.46}  \\
        60k & 16.7 & 0.43  \\
        80k & 20.4 & 0.41  \\
        \bottomrule
    \end{tabular}
    \caption{Performance of the intermediate Vikhr models with different vocabulary sizes after continued pre-training. The base model is LLaMa-2 7b. The vocabulary size selected for the final Vikhr model is underlined.}
    \label{tab:abl_vocab_size}
\end{table}

\begin{table}[t]
    \centering
    \footnotesize
    \resizebox{0.49\textwidth}{!}{
    \begin{tabular}{l|cc|c}
        \toprule
        & \textbf{PPL$\downarrow$} & \textbf{ruXNLI$\uparrow$} & \textbf{Ru-MMLU-pro$\uparrow$} \\
        \midrule
        No filt. (17b tokens) & 8.4 & 0.37 & 10.2 \\
        With filt. (11b tokens) (Ours) & \textbf{7.2} & \textbf{0.46} & \textbf{11.1} \\
        \bottomrule
    \end{tabular}
    }
    \caption{Performance metrics with and without filtration of the corpus for the continued pre-training. Perplexity is computed on the instruction dataset.}
    \label{tab:abl_filtration}
\end{table}

\begin{table}[t]
    \centering
    \footnotesize
    \begin{tabular}{l|cc}
        \toprule
        & \textbf{PPL$\downarrow$} & \textbf{ruXNLI$\uparrow$} \\
        \midrule
        No regul. & 8.1 & 0.34 \\
        With regul. (Ours) & \textbf{7.2} & \textbf{0.45} \\
        \bottomrule
    \end{tabular}
    \caption{Performance with and without KL loss regularization during continued pre-training.}
    \label{tab:abl_regul}
\end{table}

\begin{table}[t]
    \centering
    \footnotesize
    \begin{tabular}{l|cc}
    \toprule
         \textbf{Instruction Set} & \textbf{Ru-MMLU-pro} 
         \\\midrule   
         Saiga SFT & 0.21 \\
         Translated Nectar & 0.20 \\
         Ours & \textbf{0.27} \\
         \bottomrule
    \end{tabular}
    \caption{Performance of Vikhr models fine-tuned on various instruction sets. The base model is Llama-3 8b.}
    \label{tab:abl_instruct}
\end{table}

\subsection{Quality of Generated Text}

To check the performance of the model in text generation, we performed qualitative analysis of LLM outputs. Table \ref{tab:qualitative} compares several responses of Vikhr with outputs of the base model (Mistral-7B-Instruct-v0.2). As we can see from the presented examples, when Mistral generates Russian text, it often injects English words. Moreover, sometimes generated words consist of an English and a Russian token. From the second example, we see that Mistral also has issues with grammatical coherence. In the third example, LLM suggests to ``clean a fish from fur'', which illustrates the lack of understanding of word meanings in the Russian language. We also note that Mistral tends to answer in English even when the input prompt is in Russian. These issues appear very often and make the base model useless for generation of Russian texts in practical scenarios. On the considered examples, Vikhr does not demonstrate any of these problems. Texts generated by Vikhr are grammatically coherent and correct.

\subsection{Ablation Studies}

We conducted several ablation studies to demonstrate the effects of various features of our model translation pipeline.

\paragraph{Selection of the vocabulary size.} Table \ref{tab:abl_vocab_size} presents the performance of intermediate Vikhr models with different vocabulary sizes after the stage of continued pre-training. We measure perplexity of the LLM on the instruction set and the performance on the ruXNLI task \cite{conneau2018xnli}. As we can see, perplexity increases with the vocabulary size, indicating some degradation. However, the performance on the ruXNLI dataset is not monotonic. While the results for the largest vocabulary size are lower than those for the smallest, we observe a performance peak at a vocabulary size of 40k tokens. We selected this size as it offers improvements in the final task with only a slight increase in perplexity.

\paragraph{Effect of filtration of the corpus for continued pre-training} is illustrated in Table \ref{tab:abl_filtration}. As we can see, despite reducing the size of the data, performing continued pre-training on the filtered corpus results in a model with lower perplexity and substantially better scores in both considered end tasks: Ru-MMLU-pro and ruXNLI. This again highlights the importance of data quality for constructing good LLMs.

\paragraph{Effect of loss regularization in continued pre-training} is illustrated in Table \ref{tab:abl_regul}. The results show that the KL regularization introduced in our pipeline slightly reduces perplexity and substantially increases the model performance in the ruXNLI task. This shows that continued pre-training on its own might deteriorate LLM reasoning capabilities, and proper regularization helps prevent the catastrophic forgetting.

\paragraph{Effect of fine-tuning on various instruction sets} is illustrated in Table \ref{tab:abl_instruct}. The results demonstrate that fine-tuning on our instruction set gives a big boost in performance on NLU tasks compared to Saiga and translated Nectar.

\section{Conclusion}

We have presented Vikhr -- a new open-source instruction-following bilingual LLM oriented on the Russian language. To create Vikhr, we developed a comprehensive pipeline for adapting English-oriented LLMs to other languages. The pipeline includes the adaptation of the tokenizer vocabulary, continued pre-training of the entire model, and instruction tuning. We have also constructed a new dataset for instruction tuning by expanding the Saiga dataset with automatically translated and cleaned English instruction datasets. Our extensive work enabled Vikhr to achieve high quality of generation while maintaining computational efficiency. 

We hope that the developed cross-lingual adaptation pipeline and the published models will foster the research on LLMs and enhance the diversity of languages incorporated into research agendas. 

In the future work, we plan to release in the open-source new better versions of Vikhr. At the moment, our best publicly available model is Vikhr-Nemo-12B-Instruct\footnote{\url{https://huggingface.co/Vikhrmodels/Vikhr-Nemo-12B-Instruct-R-21-09-24}} based on Mistral NeMo. We also plan to perform cross-lingual adaption of LLMs to low-resource languages such as Belarusian, Serbian, and Kazakh.

\section*{Limitations}
We do not introduce additional restrictions on the usage of our models. However, the users must comply with the license of the base model and instruction datasets. 

We do not implement RLHF / DPO fine-tuning of Vikhr due to the lack of resources for human annotation. We expect further performance improvements from these techniques.

We do not introduce additional instruction-output pairs to facilitate LLM alignment. However, we note that the majority of the data for supervised fine-tuning of Vikhr are obtained from the ChatGPT model series, so our model partially inherits its alignment.

\section*{Ethical Considerations}

The development and deployment of Vikhr raise several ethical considerations that must be addressed to ensure its responsible use:

\begin{itemize}
    \item Bias and Fairness: For developing Vikhr, we use publicly available data. Despite efforts to train Vikhr on diverse datasets, there is a risk of inherent biases in the data which may be reflected in the model's outputs. Continued monitoring and evaluation are required to mitigate any biases, ensuring fair and unbiased performance.

    \item Misinformation: As with any LLM, Vikhr has the potential to generate misleading or incorrect information. It is crucial to establish guidelines and mechanisms for users to verify the information provided by the model, promoting critical assessment and cross-referencing with reliable sources.

    \item Misuse: Vikhr can be used for malicious purposes, such as generating harmful content, spam, or deepfakes. Implementing usage restrictions and monitoring mechanisms to detect and prevent misuse is critical to safeguard against these risks.

\end{itemize}

\section*{Acknowledgements}
We thank the reviewers for their insightful comments, which have significantly improved this paper. We would like to express our sincere gratitude to \textbf{Nikolay Kompanets} for his invaluable contributions to the fine-tuning of the small models and the verification of hypotheses. 

\bibliography{anthology,custom}
\bibliographystyle{acl_natbib}

\end{document}